\def\year{2020}\relax
\documentclass[letterpaper]{article} 
\usepackage{aaai20}  
\usepackage{times}  
\usepackage{helvet} 
\usepackage{courier}  
\usepackage[hyphens]{url}  
\usepackage{graphicx} 
\usepackage{graphicx}  
\usepackage{amssymb}
\usepackage{amsmath}
\usepackage{multirow}
\usepackage{xcolor}

\usepackage{caption}
\usepackage{subcaption}
\usepackage{algorithm}
\usepackage{algorithmic}
\newcommand{\citet}[1]{\citeauthor{#1} \shortcite{#1}}
\title{Neural Semantic Parsing in Low-Resource Settings with Back-Translation and Meta-Learning}
\author{Yibo Sun$^{1}$\thanks{This work was done during an internship at MSRA.},
Duyu Tang$^{2}$, Nan Duan$^{2}$, Yeyun Gong$^{2}$, Xiaocheng Feng$^{1}$, Bing Qin$^{1}$, Daxin Jiang$^{3}$
\\
 $^{1}$Harbin Institute of Technology, Harbin, China \\
 $^{2}$Microsoft Research Asia, Beijing, China \\
  $^{3}$Microsoft Search Technology Center Asia, Beijing, China \\
 \{ybsun, xcfeng, qinb\}@ir.hit.edu.cn ~~ \{dutang, nanduan, yegong, djiang\}@microsoft.com
 }
 
\begin{document}

\maketitle

	\begin{abstract}
		Neural semantic parsing has achieved impressive results in recent years, yet {its} success 
		relies on the availability of 
		large amounts of supervised data.
		Our goal is to learn a neural semantic parser when only prior knowledge about a limited number of simple rules is available, 
		without access to either annotated programs or execution results.
		Our approach is initialized by rules, and improved in a back-translation paradigm using generated question-program pairs from the semantic parser and the question generator. 
		A phrase table with frequent mapping patterns is automatically derived, also updated as training progresses, to measure the quality of generated instances. 
We train the model with model-agnostic meta-learning to guarantee the accuracy and stability on examples covered by rules, and meanwhile acquire the versatility to generalize well on examples uncovered by rules.
		Results on three benchmark datasets with different domains and programs show that our approach incrementally improves the accuracy.
		On WikiSQL, our best model is comparable to the SOTA system learned from denotations. 
	\end{abstract}
	
\begin{algorithm*}[t]
	\caption{Low-Resource Neural Semantic Parsing with Back-Translation and MAML}
	\label{alg:example}
	\begin{algorithmic}[1]
		\REQUIRE $Q$: a set of natural language questions. $LF$: a collection of sampled logical forms.
		\REQUIRE $r$: a rule, if satisfied, maps $q$ to $lf$. $\alpha$, $\beta$: step size hyperparameters. 	
		\STATE Apply $r$ to $Q$, obtain training data $D_0$
		\STATE Use $D_0$ to initialize $\theta_{q \rightarrow lf}$ and $\theta_{lf \rightarrow q}$
		\WHILE{not done}
		\STATE Apply $f_{q \rightarrow lf}$ to $Q$, apply $f_{lf \rightarrow q}$ to $LF$
		\STATE Use $f_{q \rightarrow lf}(Q)$ and $f_{lf \rightarrow q}(LF)$ to update the phrase table of the quality controller $qc$
		\STATE Update $\theta_{lf \rightarrow q}$ using $D_0$ and $qc(f_{q \rightarrow lf}(Q))$
		\FOR{Task $\mathcal{T}_i \in \{r=T, r=F\}$}
		\STATE Sample $\mathcal{D}_i$ from \{$D_0$, $qc(f_{q \rightarrow lf}(Q))$, $qc(f_{lf \rightarrow q}(LF))$\} for task $\mathcal{T}_i$
		\STATE Compute gradients using $\mathcal{D}_i$ and $\mathcal{L}$, update learner $\theta_i=\theta-\alpha \nabla_\theta  \mathcal{L}(f_\theta)$  
		\STATE Sample $\mathcal{D}_i'$ from \{$D_0$, $qc(f_{q \rightarrow lf}(Q))$, $qc(f_{lf \rightarrow q}(LF))$\} for the meta-update
		\STATE Compute gradients using $\mathcal{D}_i'$ and $\mathcal{L}$, update meta-learner $\theta = \theta - \beta \nabla_\theta \mathcal{L} ( f_{\theta_i})$
		\ENDFOR
		\STATE Update $\theta_{q \rightarrow lf}^{r=T}$ and $\theta_{q \rightarrow lf}^{r=F}$ with $\theta$ as initialization, respectively
		\ENDWHILE
	\end{algorithmic}
\end{algorithm*}

	\section{Introduction}
Semantic parsing aims to map natural language questions to the logical forms of their underlying meanings, which can be regarded as programs and executed to yield answers, aka denotations \cite{berant2013semantic}. 
%
In the past few years, neural network based semantic parsers have achieved promising performances \cite{liang-EtAl:2017:Long}, however, their success is limited to the setting with 
rich supervision, 
which is costly to obtain.
There have been recent \mbox{attempts} at low-resource semantic parsing, including data augmentation methods which are learned from a small number of annotated examples \cite{Guo18-1188}, and methods for adapting to unseen domains while \mbox{only} being trained on annotated examples in other domains.

This work investigates neural semantic parsing in a low-resource setting, in which case we only have our prior knowledge about a limited number of simple mapping rules, including a small amount of domain-independent word-level matching tables if necessary, 
but have no access to either annotated programs or execution results.
Our key idea is to use these rules to collect modest question-programs pairs as the starting point, and then leverage automatically generated examples to improve the accuracy and generality of the model. 
This presents three challenges including how to generate examples in an \mbox{efficient} way, how to measure the quality of generated examples which might contain errors and noise, and how to train a semantic parser that makes robust predictions for examples covered by rules and generalizes well to uncovered examples.

We address the aforementioned challenges with a framework consisting of three key components.
The \textbf{first} component is a data generator. 
It includes a neural semantic parsing model,
which maps a natural language question to a program, 
and a neural question generation model,
which maps a program to a natural language question. 
We learn these \mbox{two} \mbox{models} in a back-translation paradigm using pseudo parallel examples, inspired by its big success on unsupervised neural machine translation \cite{sennrich16-1009,lample18}.
The \textbf{second} \mbox{component} is a quality controller, which is used for filtering out noise and errors contained in the pseudo data. 
We \mbox{construct} a phrase table with frequent mapping patterns, therefore  noise and errors with low frequency can be filtered out. A similar idea has been worked as posterior regularization in neural machine translation \cite{Zhang17-1139,ren2019}.
The \textbf{third} component is a meta learner. Instead of transferring a model pretrained with examples covered by rules to the generated examples, we leverage model-agnostic meta-learning \cite{finn2017model}, an elegant meta-learning algorithm which has been successfully applied to a wide range of tasks including few-shot learning and adaptive control. 
We regard different data sources as different tasks, and 
use 
outputs of the quality controller 
for stable training.

We test our approach on three tasks with different programs, including SQL (and SQL-like) queries for both single-turn and multi-turn questions over web tables \cite{zhong2017seq2sql,iyyer-yih-chang:2017:Long}, 
and subject-predicate pairs over a large-scale knowledge graph \cite{bordesUCW15}.
The program for SQL queries for single-turn questions and subject-predicate pairs over knowledge graph is simple while the program for SQL queries for multi-turn questions have top-tier complexity among currently proposed tasks.
Results show that our approach yields large improvements over rule-based systems, 
and incorporating different strategies incrementally improves the overall performance. 
On WikiSQL, our best performing system achieves execution accuracy of 72.7\%, comparable to a strong system learned from denotations \cite{Agarwal2019LearningTG} with an accuracy of 74.8\%.

\section{Problem Statement}
We focus on the task of executive semantic parsing.
The goal is to map a natural language question/utterance $q$ to a logical form/program $lf$, which can be executed over a world $Wld$ to obtain the correct answer $a$. 

We consider three tasks. 
The \textbf{first} task is single-turn table-based semantic parsing, in which case $q$ is a self-contained question, $lf$ is a SQL query in the form of \textit{``SELECT\ agg\ col$_1$\ WHERE\ col$_2$\ =\ val$_2$\ AND\ ...''}, and $Wld$ is a web table consisting of multiple rows and columns. We use WikiSQL \cite{zhong2017seq2sql} as the testbed for this task.
The \textbf{second} task is multi-turn table-based semantic parsing. Compared to the first task, $q$ could be a follow-up question, the meaning of which depends on the conversation history. Accordingly, $lf$ in this task supports additional operations that copy previous turn $lf$ to the current turn. We use SequentialQA  \cite{iyyer-yih-chang:2017:Long} for evaluation.
In the \textbf{third} task, we change $Wld$ to a large-scale knowledge-graph (i.e. Freebase) and consider knowledge-based question answering for single-turn questions. We use SimpleQuestions \cite{bordesUCW15} as the testbed, where the $lf$ is in the form of a simple $\lambda$-calculus like $\lambda x.predicate(subject,x)$, and the generation of $lf$ is equivalent to the prediction of the predicate and the subject entity.

We study the problem in a low-resource setting.
In the training process, we don't have annotated logical forms $lf$ or execution results $a$.
Instead, we have  a collection of natural language \mbox{questions} for the task, a limited number of simple mapping rules based on our prior knowledge about the task, and may also have a small \mbox{amount} of domain-independent word-level matching tables if necessary.
These rules are not perfect, with low coverage, and can even be incorrect for some \mbox{situations}.
For instance, when predicting a SQL command in the first task, we have a prior knowledge that (1) WHERE values potentially have co-occurring words with table cells; (2) the words ``\textit{more}'' and ``\textit{greater}'' tend to be mapped to WHERE operator ``$>$''; (3) within a WHERE clause, header and cell should be in the same column; and (4) the word ``\textit{average}'' tends to be mapped to aggregator ``\textit{avg}''.
Similarly, when predicting a $\lambda$-calculus in the third task, the entity name might be present in the question, and among all the predicates connected to the entity, the predicate with maximum number of co-occurred words might be correct.
We would like to study to what extent our model can achieve if we use rules as the starting point.
\begin{figure*}[t]
	\centering
	\includegraphics[width=\textwidth]{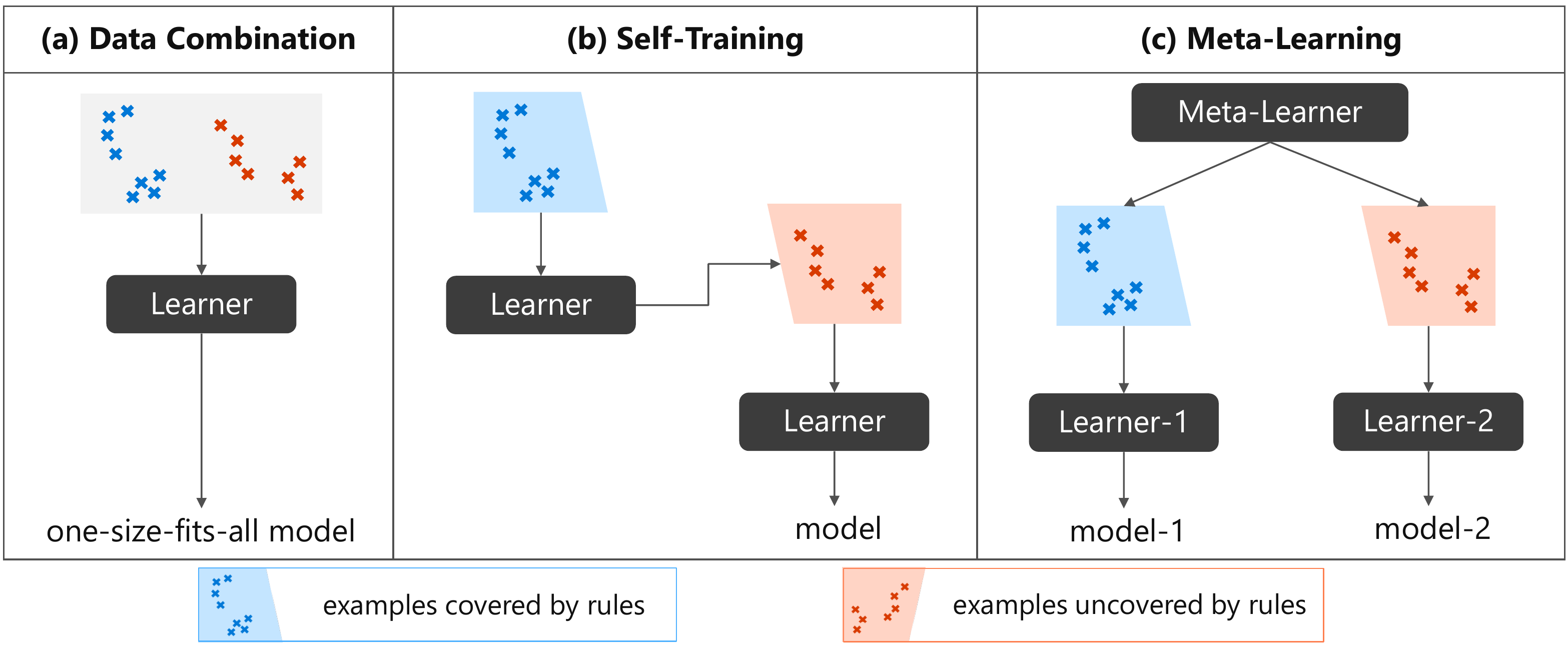}
	\caption{An illustration of the difference between (a) data combination which learns a monolithic, one-size-fits-all model, (b) self-training which learns from predictions which the model produce and (c) meta-learning that reuse the acquired ability to learn. }
	\label{fig:meta-self}
\end{figure*}
\section{Learning Algorithm}
We describe our approach for low-resource neural semantic parsing in this section. 

We propose to train a neural semantic parser using back-translation and meta-learning. 
The learning process is summarized in Algorithm \ref{alg:example}. We describe the three components in this section, namely back-translation, quality control, and meta-learning.

\subsection{Back-Translation}
Following the back-translation paradigm \cite{sennrich16-1009,lample18}, we have a semantic parser, which maps a natural language question $q$ to a logical form $lf$, and a question generator, which maps $lf$ to $q$. The semantic parser works for the primary task, and the question generator mainly works for generating pseudo datapoints.
We start the training process by applying the rule $r$ to a set of natural language \mbox{questions} $Q$. 
The resulting dataset is considered as the training {data} to initialize both the semantic parser and the question generator. 
\mbox{Afterwards}, both \mbox{models} are improved following the back-translation protocol that target sequences should follow the real data distribution, yet source sequences can be generated with noises. This is based on the consideration that 
in an encoder-decoder model, the decoder is more sensitive to the data distribution than the encoder.
We use datapoints from both models to train the semantic parser because a logical form is structural which follows a grammar, whose distribution is similar to the ground truth.

\subsection{Quality Controller}
Directly using generated datapoints as supervised training data is not desirable because those generated datapoints contain noises or errors.
To address this, we follow the application of posterior regularization in neural machine translation \cite{Zhang17-1139}, and implement a dictionary-based discriminator which is used to measure the quality of a pseudo data.
The basic idea is that although these generated datapoints are not perfect, the frequent patterns of the mapping between a phrase in $q$ to a token in $lf$ are helpful in filtering out noise  in the generated data with low frequency \cite{ren2019}.
There are multiple ways to collect the phrase table information, such as using statistical phrase-level alignment algorithms like Giza++ or directly counting the co-occurrence of any question word and logical form token. We use the latter one in this work. Further details are described in the appendix.

\subsection{Meta-Learning}
A simple way to update the semantic parser is to merge the datapoints in hand and train a one-size-fits-all model \cite{Guo18-1188}.
However, this will hurt model's stability on examples covered by rules, and examples of the same task may vary widely \cite{huang18}. Dealing with different types of examples requires the model to possess different abilities. For example, tackling examples uncovered by rules in WikiSQL requires the model to have the additional ability to map a column name to a totally different utterance, such as ``\textit{country}'' to ``\textit{nation}''. 
Another simple solution is self-training \cite{mcclosky2006effective}. One can train a model with examples covered by rules, and use the model as a teacher to make predictions on examples uncovered by rules and update the model on these \mbox{predictions}.
However, self-training is somewhat tautological because the model is learned to make predictions which it already can produce.

We learn the semantic parser with meta-learning, regarding learning from examples covered by rules or uncovered by rules as two (pseudo) tasks.
Compared to the aforementioned strategies, the advantage of exploring meta-learning here is two-fold.
\underline{First}, we learn a specific model for each task, which provides guarantees about its stability on examples covered by rules.
In the \textbf{test phase}, we can use the rule to detect which task an example belongs to, and use the corresponding task-specific model to make predictions. When dealing with examples covered by rules, we can either directly use rules to make predictions or use the updated model, depending on the accuracy of the learned model on the examples covered by rules on development set. 
\underline{Second}, latent patterns of examples may vary widely in terms of whether or not they are covered by rules. 
Meta-learning is more desirable in this situation because it \mbox{learns} the model's ability to learn, improving model's versatility rather than mapping the latent patterns learned from datapoints in one distribution to datapoints in another distribution by force.
Figure \ref{fig:meta-self} is an illustration of data combination, self-training, and meta-learning.

\begin{table*}[t]
	\centering
	\begin{tabular}{l|c|ccc}
		\hline

		{Methods} & Supervision& {Rule Covered} & {Rule Uncovered}& {Overall}\\
		\hline
		\citet{dong18coarsetofine} & logical form& -& -& 78.5\% \\
 		\citet{shi2018incsql} & logical form& -& -& 87.1\%\\
 		\citet{liangNIPS2018_8204}& denotation& -& -& 72.4\% \\
 		\citet{Agarwal2019LearningTG}& denotation& -& -& 74.8\% \\
 		
		\hline    
		Base (full supervision) & logical form & 85.0\% & 71.7\% & 82.3\% \\

		Rule & {rule}& 77.6\%  & 0.0\% & 61.8\% \\
		Base (trained on rule-covered data) & rule& 75.9\% & 48.4\% & 70.3\% \\
		Base + Self Training & rule& 75.6\% &49.3\% & 70.3\% \\
		Base + Question Generation& rule& 77.7\% &51.9\% & 72.5\% \\
		Base + BT & rule& 77.6\% &51.6\% & 72.3\% \\
		Base + BT + QC& rule& 77.8\% &52.1\% & 72.6\% \\
		Base + BT + QC + MAML& rule& 77.9\%& 52.1\%& 72.7\%\\

		\hline
	\end{tabular}
	
	\caption{Results on WikiSQL testset. BT stands for back-translation. QC stands for quality control.}
	\label{table:wikisql-results}
\end{table*}

Meta-learning includes two optimizations: {the learner} that \mbox{learns} new tasks, and the {meta-learner} that trains the learner.
In this work, the meta-learner is optimized by finding a good initialization that is highly adaptable. 
Specifically, we use model-agnostic meta-learning, \mbox{MAML} \cite{finn2017model}, a powerful meta-learning algorithm with desirable properties including introducing no additional \mbox{parameters} and making no assumptions of the form of the model. 
In \mbox{MAML}, task-specific parameter $\theta_i$ is initialized by $\theta$, and updated using gradient decent based on the loss function $\mathcal{L}_i$ of task $\mathcal{T}_i$. 
In this work, the loss functions of two tasks are the same. 
The updated parameter $\theta_i$ is then used to calculate the model's performance across tasks to update the parameter $\theta$.
In this work, following the practical \mbox{suggestions} given by \citet{antoniou2019train}, we update $\theta$ in the \textit{inner-loop} and regard  the outputs of the quality controller as the input of both tasks.

If we only have examples covered by rules, such as those used in the initialization phase, meta-learning learns to learn a good initial parameter that is evaluated by its usefulness on the examples from the same distribution. 
In the training phase, datapoints from both tasks are generated, and meta-learning learns to learn an initialization parameter which can be quickly and efficiently adapted to examples from both tasks. 

\section{Experiment}
We conduct experiments on three tasks to test our approach, including generating SQL (or SQL-like) queries for both single-turn and multi-turn questions over web tables \cite{zhong2017seq2sql,iyyer-yih-chang:2017:Long}, 
and predicting subject-predicate pairs over a knowledge graph \cite{bordesUCW15}. 
We describe task definition, base models, experiments settings and empirical results for each task, respectively.

\subsection{Table-Based Semantic Parsing}
\label{section:wikisql}
\paragraph{Task and Dataset}
Given a natural language 
$q$ and a table $t$ with $n$ columns and $m$ rows as the input, the task is to output a SQL query $y$, which could be executed on table $t$ to yield the correct answer of $q$.
We conduct experiments on WikiSQL \cite{zhong2017seq2sql}, which provides 87,726 annotated question-SQL pairs over 26,375 web tables. In this work, we do not use either SQL queries or answers in the training process. 
We use execution accuracy as the evaluation metric, which measures the percentage of generated SQL queries that result in the correct answer. 

 \begin{table}[h]
 	\centering
 	\begin{tabular}{c|p{4.75cm}}
 		\hline
 		SQL Token & NL Word\\
 		\hline
 		$sum$  & sum\\
 		$count$  & how many, total number\\
 		$max$ & maximum\\
 		$min$ & minimum\\
 		$avg$ & average\\
 		\hline
 		$>$ & more, greater, higher, taller, longer, older, larger, after\\
 		$<$ & less, smaller, lower, fewer, nearer, shorter, before\\
 		\hline
 	\end{tabular}
 	\caption{Token-level dictionary for aggregators (upper group) and operators (lower group) in WikiSQL.}
 	\label{table:wikisql-rule-keywords}
 \end{table}

\paragraph{Rules}
We describe our rules for WikiSQL here.
We first detect WHERE values, which exactly match to table cells.
After that, if a cell appears at more than one column, we choose the column name with more overlapped words with the question, with a constraint that the number of co-occurred words is larger than 1.
By default, a WHERE operator is $=$, except for the case that surrounding words of a value contain keywords for $>$ and $<$. 
Then, we deal with the SELECT column, which has the largest number of co-occurred words and cannot be same with any WHERE column.
By default, the SELECT AGG is NONE, except for matching to any keywords in Table \ref{table:wikisql-rule-keywords}.
The coverage of our rule on training set is 78.4\%, with  execution accuracy of 77.9\%.

\begin{table*}[t]
	\centering
	\begin{tabular}{l|c|ccc}
		\hline
		Methods& Supervision& {Rule Covered} & {Rule Uncovered} & {Overall}\\
		\hline    
		Base (full supervision) & logical form & 81.8\% & 67.6\% & 72.4\%  \\
		Rule & rule& 76.5\% & 0.0\%  & 25.7\%  \\
		Base (trained on rule-covered data) & rule& 74.2\% & 34.8\% & 48.0\%  \\
		Base + Self Training &rule & 74.4\% & 46.5\% & 55.9\%  \\
		Base + Question Generation & rule& 73.9\% & 42.1\% & 52.8\%  \\
		Base + BT &rule & 75.5\% & 47.7\% & 57.1\%  \\
		Base + BT + QC & rule& 75.3\% & 48.6\% & 57.6\%  \\
		Base + BT + QC + MAML & rule& 76.8\% & 48.6\% & 58.1\%  \\
		\hline
	\end{tabular}

		\caption{Results on SimpleQuestions testset. BT stands for back-translation. QC stands for quality control.}
	\label{table:simpleq-results}
\end{table*}

\paragraph{Base Model}
We implement a neural network modular approach as the base model,
%
which includes different modules to predict different SQL constituents.
This approach is based on the understanding of the SQL grammar in WikiSQL, namely ``\textit{SELECT \$agg \$column  WHERE \$column \$op \$value (AND \$column \$op \$value)*}'', where tokens starting with ``\textit{\$}'' are the slots to be predicted \cite{xu2017sqlnet}. In practice, modular approaches typically achieve higher accuracy than end-to-end learning approach. 
Specifically, at the first step we implement a sequential labeling module to detect 
WHERE values and link them to table \mbox{cells}.
Advantages of starting from WHERE values include that WHERE values are less ambiguous compared to other slots, and that the number of WHERE clauses can be  
naturally detected.
After that, for each WHERE value, we use the preceding and following contexts in the question to predict its WHERE column and the WHERE operator through two unidirectional LSTM.
Column attention \cite{xu2017sqlnet} is used 
for predicting a particular column.
Similar LSTM-based classifiers are used to predict SELECT column and \mbox{SELECT} aggregator.
\begin{figure}[h]
	\centering
	
	\begin{subfigure}{.22\textwidth}
		\centering
		\includegraphics[width=\linewidth]{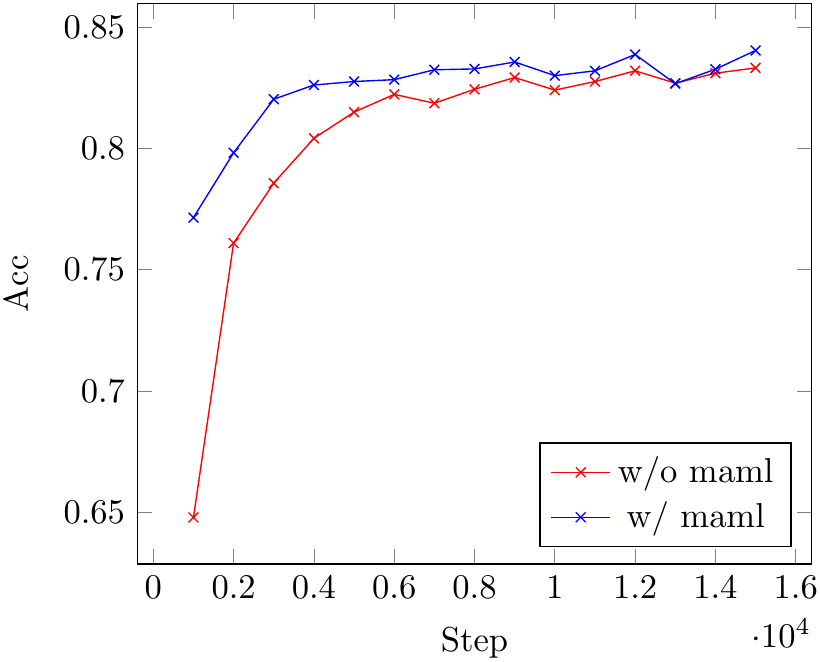}
		\caption{Rule-covered}
		\label{fig:sub1}
	\end{subfigure}%
	\begin{subfigure}{.22\textwidth}
		\centering
		\includegraphics[width=\linewidth]{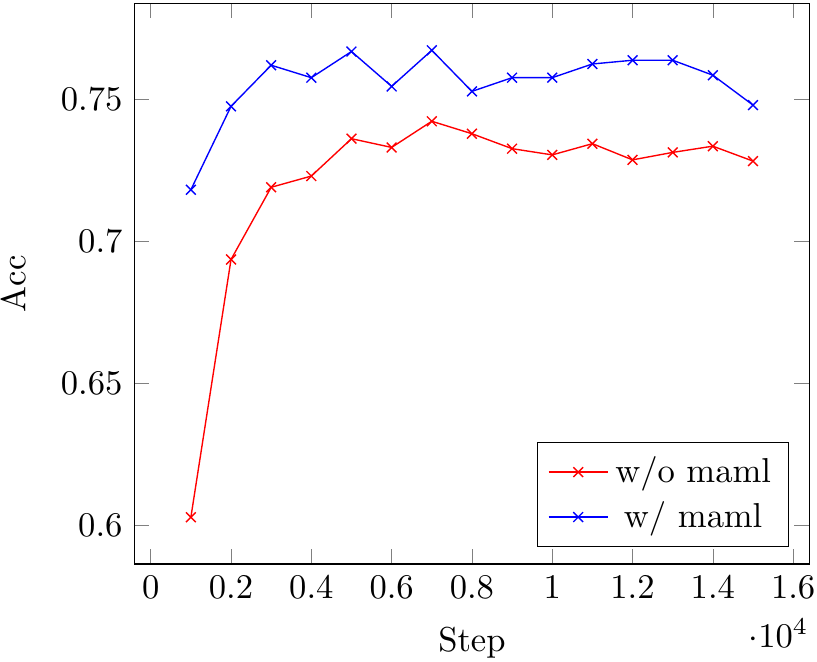}
		\caption{Rule-uncovered}
		\label{fig:sub2}
	\end{subfigure}
	\caption{Learning curve of the WHERE column prediction model on WikiSQL devset. }
	\label{fig:curve}
\end{figure}

\paragraph{Settings}
According to whether the training data can be processed by our rules, we divide it into two parts: rule covered part and rule uncovered part.
For the rule covered part we could get rule covered training data using our rules.
For the rule uncovered part we could also get training data using the trained Base model we have, we refer to these data as self-inference training data.
Furthermore, we could get more training data by back translation, we refer to these data as question-generation training data.
For all the settings, the Base Model is initialized  with rule covered training data.
In Base + Self Training Method, we finetune the Base model with self-inference training data.
In Base + Question Generation Method, we use question-generation training data to finetune our model.
In Base + BT Method, we use both self-inference and question-generation data to finetune our model.
In Base + BT + QC, we add our quality controller. In Base + BT + QC + MAML, we further add meta-learning.
\paragraph{Results and Analysis}
Results are given in Table \ref{table:wikisql-results}. We can see that back-translation, quality control and \mbox{MAML} incrementally improves the accuracy.
Question generation is better than self-training here because the logical form in WikiSQL is relatively simple, so the distribution of the sampled logical forms is similar to the original one.
In the back-translation setting, generated examples come from both self-training and the question generation model.
The model performs better than rules on rule-covered examples, and improves the accuracy on uncovered examples.\
Figure \ref{fig:curve} shows the learning curves of the COLUMN prediction model with or without using MAML. The model using MAML has a better starting point during training, which reflects the effectiveness of the pre-trained parameter.

\subsection{Knowledge-Based Question Answering}
We test our approach on question answering over another genre of \mbox{environment}: knowledge graph consisting of subject-relation-object triples.

\paragraph{Task and Dataset}
Given a natural language question and a knowledge graph, the task aims to correctly answer the question with evidences from the knowledge graph.
We do our study on SimpleQuestions \cite{bordesUCW15}, which includes 108,442 simple questions, each of which is accompanied by a subject-relation-object triple.
Questions are constructed in a way that subject and relation are mentioned in the question, and that object is the answer.
The task requires predicting the entityId and the relation involved in the question.  


\paragraph{Rulse}
Our rule for KBQA is simple without using a curated mapping dictionary.
First, we detect an entity from the question using strict string matching, with the constraint that only one entity from the KB has the same surface string and that the question contains only one entity.
After that, we get the connected relations of the detected entity, and assign the relation as the one with maximum number of co-occurred words.
The coverage of our rule on training set is 16.0\%, with an accuracy of 97.3\% for relation prediction.

\begin{table*}[t]
	\centering
	\begin{tabular}{l|c|c|c|c}
		\hline
		Methods & Supervision & Rule Covered & Rule Uncovered & Overall \\
		\hline
\citet{pasupat-liang:2015:ACL-IJCNLP} & denotation&- & -& 33.2\%  \\
\citet{neelakantan2016learning} & denotation&- & -& 40.2\%  \\
\citet{iyyer-yih-chang:2017:Long} &denotation &- & -& 44.7\%  \\
\citet{misra2018policy} &denotation &- & -& 49.7\% \\
		\hline
		Base (full supervision) & denotation& 50.6\% & 33.5\% & 45.9\% \\
		Rule & rule& 51.2\% & 0.0\% & 37.2\%  \\
		Base (trained on rule-covered data) & rule& 43.7\% & 9.3\% & 34.3\%  \\
		Base + Self Training & rule& 42.3\% & 10.7\% & 33.7\%  \\
		Base + Question Generation & rule& 34.7\% & 9.1\% & 27.7\%  \\
		Base + BT &rule & 39.2\%& 10.1\% & 31.3\%  \\
		Base + BT + QC & rule& 41.0\% & 10.1\% & 32.6\% \\
 		Base + BT + QC + MAML& rule& 41.2\%& 9.7\% & 32.7\% \\
 		Base + BT (w/o QG) + MAML &rule & 43.7\%& 11.3\% & 34.6\%  \\
		\hline 
	\end{tabular}
	
\caption{Results on SequentialQA testset. BT stands for back-translation. QC stands for quality control.}
	\label{table:sqa-results}
\end{table*}

\paragraph{Base Model}
We follow \citet{petrochuk18simpleqsolved}, and implement a KBQA pipeline consisting of three modules in this work. 
At the first step, we use a sequence labeling model, i.e. LSTM-CRF, to detect entity mention words in the question. 
After that, we use an entity linking model with BM25 built on Elasticsearch.
Top-K ranked similar entities are retrieved as candidate list.
Then, we get all the relations connected to entities in the candidate list as candidate \mbox{relations}, and use a relation prediction model, which is based on Match-LSTM \cite{wangJ16aMatchlstm}, to predict the relation. 
Finally, from all the entities connected to the predicted relation, we choose the one with highest BM25 score as the predicted entity. We use FB2M as the KB, which includes about 2 million triples. 

\paragraph{Settings}
The settings are the same as those described in table-based semantic parsing.

\begin{table}[h]
 	\centering
 	\begin{tabular}{c|p{4.75cm}}
 		\hline
 		SQL Token & NL Word\\
 		\hline
 		$\geq$  & no less than, greater or equal, at least\\
 		$\leq$  & no more than, less or equal, at most\\
 		\hline
 		$followup$ & of those, which ones, which one\\
 		\hline
 		$NEG$ & do not, does not, did not, have not,
 		has not, had not, was not, were not, is not, are not,
 		not have, not has\\
 		\hline
 	\end{tabular}
 	\caption{Token-level dictionary used for additional actions in SequentialQA.}
 	\label{table:sqa-rule-keywords}
 \end{table}
 
\paragraph{Results and Analysis}

Results are given in Table \ref{table:simpleq-results}, which are consistent with the numbers in WikiSQL. Using back-translation, quality control and \mbox{MAML} incrementally improves the accuracy, and our approach generalizes well to rule-uncovered examples.

\subsection{Conversational Table-Based Semantic Parsing}
\label{section:sqa}
We consider the task of conversational table-based semantic parsing in this part. 
Compared to single-turn table-based semantic parsing as described in subsection \ref{section:wikisql}, the meaning of a natural language may also depends on questions of past \mbox{turns}, which is the common ellipsis and co-reference phenomena in conversational agents. 

\paragraph{Task and Dataset}
Given a natural language question at the current turn, a web table, and previous turn questions in a conversation as the input, the task aims to generate a program (i.e. logical form), which can be executed on the table to obtain the correct answer of the current turn question.

We conduct experiments on SequentialQA \cite{iyyer-yih-chang:2017:Long} which is derived from the WikiTableQuestions dataset\cite{pasupat-liang:2015:ACL-IJCNLP}. It contains 6,066 question sequences covering 17,553 question-answer pairs.
Each sequence includes 2.9 natural language questions on average.
Different from WikiSQL which provides the correct logical form for each question, SequentialQA only annotates the correct answer. This dataset is also harder than the previous two, since it requires complex, highly compositional logical forms to get the answer.
Existing approaches are evaluated by question answering accuracy, 
which measures whether the predicted answer is correct or not.

\paragraph{Rules}
The pipeline of rules in SequentialQA is similar to that of WikiSQL.
Compared to the grammar of WikiSQL, the grammar of SequentialQA has additional actions including copying the previous-turn logical form, no greater than, no more than, and negation.
Table \ref{table:sqa-rule-keywords} shows the additional word-level mapping table used in SequentialQA.
The coverage of our rule on training set is 75.5\%, with an accuracy of 38.5\%.

\paragraph{Base Model}
We implement a modular approach on top of a grammar of derivation rules (actions) as the base model. 
Similar to \citet{iyyer-yih-chang:2017:Long}, our grammar consists of predefined actions used for predicting SELECT \mbox{column}, WHERE column, WHERE operator, WHERE value, and determining whether it is required to copy the entire action sequence of the previous turn questions.
After encoding a question and previous turn questions into vectors, we first use a controller module to predict an action sequence consisting of slots, and then use specific modules to predict the argument of each slot.
Similar to \citet{iyyer-yih-chang:2017:Long}, we use a recurrent structure as the backbone of each module and use the softmax layer for making prediction. 
\paragraph{Settings}
The settings are the same as those described in table-based semantic parsing.
\paragraph{Results and Analysis}
From Table \ref{table:sqa-results}, we can see that question generation does not work well on this task. This is because the difficulty in generating sequential questions and complex target logical forms.
Applying MAML to examples not coming from question generation performs best.
We leave contextual question generation as a future work.

\section{Related Work}
\paragraph{Semantic Parsing}
Semantic parsing aims to transform a natural language utterance to a program/logical form, which is executable on an environment, or a world, to obtain the result. 
There are a variety of semantic parsing tasks with different types of {environments} (including large-scale knowledge base \cite{berant2013semantic}, web table \cite{pasupat-liang:2015:ACL-IJCNLP}, image, 3D \mbox{environment}, etc.), and different types of programs (such as $\lambda$-calculus, dependency-based compositional semantics, SQL query \cite{zhong2017seq2sql}, Bash command, source code, etc.). 
The majority of existing works study the problem in a supervised or weak-supervised setting, in which case either annotated programs or execution results are available during training. 
Earlier works on unsupervised semantic parsing do not ground to an environment such a KB or a web table \cite{poon2009unsupervised}.
\citet{goldwasser2011confidence} align words to predicates, then do compositions and self training by iteratively adding self-annotated examples.
\citet{krishnamurthy2012weakly} identify relation instances from KB and produce parses that syntactically agree with the dependency parses.
\citet{reddy2014large} parse sentences with CCG, and map results to ungrounded and grounded graph by regarding semantic parsing as graph matching.
There are recent attempts at combining a limited number of supervised datapoints and artificially generated programs \cite{Guo18-1188} with generative models.

More recently, \citet{cheng2018building} share the same motivation with this work that learns semantic parser from templates. We differ from them in two aspects. First, we only need domain-general rules but they also require domain-specific lexicon containing mapping from natural language expressions to database predicates/entities.
Second, our method innovates the coupling of back-translation and model-agnostic meta-learning.

\paragraph{Meta-Learning}
Meta-learning, also known as learning to learn, is  one of the potential techniques for enabling an artificial agent to mimic a human's ability in using past experience to quickly adapt to unseen situations.
Existing approaches largely \mbox{fall} into two categories.
In the first category, neural network models are trained to learn from datapoints that are passed in. 
Different neural architectures including LSTM, convolution, or memory-augmented ones have been explored.
\mbox{Methods} in this category either use additional datapoints as a part of the input to predict the label for the new example, or optimizes the meta-learning based on the performance of the updated parameter of the learner \cite{finn2017model}.
The second category aims to learn a metric space, where examples with the same class are close with each other.
Notable works include Matching Networks and Prototypical Networks, both of which are tested on low-resource image classification.
Matching networks use attention mechanisms to use the similarity between datapoints to predict the label of a test example.
Prototypical Networks approach considers the representation of each class, which is obtained by averaging representation of examples belonging to that class.
We use MAML \cite{finn2017model}, which has been recently used in low-resource multilingual neural machine translation and sequence-to-SQL generation \cite{huang18} in supervised settings.

\paragraph{Back-Translation}
Back-translation is introduced for improving neural machine translation (NMT) by injecting monolingual data \cite{sennrich16-1009}, and is the key contributor to drive the recent success on unsupervised NMT \cite{lample18}.
The training process involves a generative source-to-target translator $f_{s\rightarrow t}$ and a generative target-to-source translator $f_{t\rightarrow s}$. Because generative models are sensitive to the target distribution while tolerant to the noises from input, target sequences typically come from the real distribution during NMT training.
Namely, examples generated by $f_{s\rightarrow t}$ are used to train $f_{t\rightarrow s}$, and examples generated by $f_{t\rightarrow s}$ are used to train $f_{s\rightarrow t}$.
In semantic parsing, we do not regard sampled programs as noisy because they follow a certain grammar which can be guaranteed to be correctly executed.
\section{Conclusion and Future Directions}
We present an approach to learn neural semantic parser from simple domain-independent rules, instead of annotated logical forms or \mbox{denotations}.
Our approach starts from examples covered by rules, which are used to initialize a semantic parser and a question generator in a back-translation paradigm.
Generated examples are measured and filtered based on statistic analysis, and then used with model-agnostic meta-learning, which guarantees model's accuracy and stability on rule-covered examples, and acquires the versatility to generalize well on rule-uncovered examples.
We conduct experiments on three datasets for table-based and knowledge-based question answering tasks.
Results show that incorporating different strategies incrementally improves the performance.
Our best model on WikiSQL
  achieves comparable accuracy to the system learned from denotation.
In the future, we plan to 
focus on more complex logical forms.
\bibliographystyle{./aaai}
\bibliography{acl2018}
\end{document}